\pdfoutput=1
\documentclass[11pt]{article}
\usepackage{tabularx}
\usepackage{multirow}
\usepackage[utf8]{inputenc} % allow utf-8 input
\usepackage[T1]{fontenc}    % use 8-bit T1 fonts
\usepackage{hyperref}       % hyperlinks
\usepackage{url}            % simple URL typesetting
\usepackage{booktabs}       % professional-quality tables
\usepackage{amsfonts}       % blackboard math symbols
\usepackage{nicefrac}       % compact symbols for 1/2, etc.
\usepackage{microtype}      % microtypography
\usepackage{xcolor}         % colors
\usepackage{caption}
\usepackage{microtype}
\usepackage{comment}
\usepackage{graphicx}
\usepackage{subfigure}
\usepackage{multirow}
\usepackage{array}
\usepackage{makecell}
\usepackage{float}
\usepackage{amsmath}
\usepackage{amssymb}
\usepackage{subfiles}
\usepackage{algorithm}

\usepackage{tabularx}% <-- added
\usepackage[export]{adjustbox}%
\usepackage{EMNLP2022}

\usepackage{times}
\usepackage{latexsym}

\usepackage[T1]{fontenc}
\usepackage[utf8]{inputenc}

\usepackage{microtype}

\usepackage{inconsolata}
\usepackage{amsmath}
\usepackage{graphicx}
\title{Understanding and Improving Knowledge Distillation for Quantization-Aware Training of Large Transformer Encoders}

\author{Minsoo Kim\textsuperscript{1}, Sihwa Lee\textsuperscript{2}, Sukjin Hong\textsuperscript{3}, Du-Seong Chang\textsuperscript{3}, and Jungwook Choi\textsuperscript{1,2}\thanks{\,\,\,Corresponding Author}  \\
        \normalsize{\textsuperscript{1}Department of Electronic Engineering, Hanyang University} \\
        \normalsize{\textsuperscript{2}Department of Artificial Intelligence, Hanyang University} \\
        \normalsize{Seoul, Republic of Korea} \\
        \small{\texttt{\{minsoo2333, macto94, choij\}@hanyang.ac.kr}},\\
        \normalsize{\textsuperscript{3}KT, Seoul, Republic of Korea}\\
        \small{\texttt{\{sukjin.hong, dschang\}@kt.com}}\\
} 
        
\begin{document}
\maketitle
\begin{abstract}
Knowledge distillation (KD) has been a ubiquitous method for model compression to strengthen the capability of a lightweight model with the transferred knowledge from the teacher. In particular, KD has been employed in quantization-aware training (QAT) of Transformer encoders like BERT to improve the accuracy of the student model with the reduced-precision weight parameters. However, little is understood about which of the various KD approaches best fits the QAT of Transformers. In this work, we provide an in-depth analysis of the mechanism of KD on attention recovery of quantized large Transformers. In particular, we reveal that the previously adopted MSE loss on the attention score is insufficient for recovering the self-attention information. Therefore, we propose two KD methods; attention-map and attention-output losses. Furthermore, we explore the unification of both losses to address task-dependent preference between attention-map and output losses. The experimental results on various Transformer encoder models demonstrate that the proposed KD methods achieve state-of-the-art accuracy for QAT with sub-2-bit weight quantization.
\end{abstract}

% --------------------------------------------------------------------------- %
% Introduction
% --------------------------------------------------------------------------- %
\section{Introduction}
\label{sec:introduction}

%% Definition of KD. Variants methods of KD (logits, features). KD is widely used in model compression of BERT
Knowledge distillation (KD)~\cite{kd} is a transfer learning framework to pass on knowledge of a large model (the teacher) to a lightweight model (the student). Numerous KD methods have been developed regarding the source of knowledge and distillation objectives. In many cases, Kullback-Leibler divergence (KL-Div) is used as a default distillation objective to match the soft labels of the teacher and the student~\cite{kd,distilbert}. But further studies on KD suggest that internal representations also convey the intermediate knowledge of the teacher~\cite{pkd,aguilar2020knowledge}. Thus minimizing the distance (e.g., mean squared error, MSE) of hidden state knowledge (HSK) of layers between the teacher and the student has also been proposed~\cite{pkd, marginal}. KD becomes an essential model compression technique for efficiently deploying large-scale Transformer-based language models. For example, a popular Transformer encoder model, BERT, contains hundreds of millions of parameters, incurring profound memory and computation overhead \cite{devlin2018bert}. These large-scale models require extreme compression to reduce the model footprint by 10 to 100 times. Therefore, extensive studies have been accomplished to distill efficient student models ~\cite{distilbert,pkd,tinybert,minilm,wang2021minilmv2}, but their focus is limited to achieving fewer parameters.

%% KD for QAT of BERT. KD is essential for successful QAT of BERT, but there is lack of knowledge
Quantization-aware training (QAT) stands out for its recent success in reducing not only the memory requirements but also the computational complexity of Transformer models \cite{bhandare2019efficient,q8bert,ibert}. Although QAT reflects quantization errors during the forward pass computation of stochastic gradient descent to train a model more robust to quantization errors, quantizing weight parameters of Transformers to a precision lower than 2-bits degrades the accuracy. Therefore, many recent QAT techniques employed the KD framework to distill the capability of the full-precision teacher to the student model with reduced-precision parameters~\cite{ternarybert, binarybert, jin2021kdlsq, li2022dq}. However, little is understood about which of the various KD approaches best fits the QAT of large Transformers. Without careful justification, most prior works adopted the layer-wise distillation of the attention score and the Transformer output with the MSE loss in addition to the basic KL-Div loss on the model output. Therefore, it is unclear if such KD setting is the most helpful for QAT on large-scale Transformer encoders like BERT-Large.

%% This work: Investigate the role of KD on QAT
In this work, we provide an in-depth analysis of KD on attention recovery for QAT of Transformers in terms of the knowledge sources and the objectives. We first reveal that all-layer KD of the intermediate Transformer layer is essential for QAT, in contrast to the KD-based model compression. 
In the case of BERT-Base, we further discover that the KL-Div-based KD on attention-map (called attention-map loss) outperforms the prior KD technique that takes MSE loss on the attention score. However, the attention-map loss is insufficient for the large Transformer encoders since weight quantization disrupts attention propagation for specific NLP tasks when there are many layers. Therefore, we devise an insightful KD, MSE loss on attention output (called attention-output loss), and help preserve attention recovery along with many layers. The proposed attention-map and output losses and their combination are evaluated on various Transformer encoder models (BERT-Base/Large and a BERT-like Korean language model (ULM). The experimental results demonstrate that the proposed KD methods significantly boost the model accuracy surpassing the state-of-the-art for QAT with aggressive sub-2-bit weight quantization.

%% Contribution
We summarize our contributions as follows:
\begin{itemize}
\item We improve the prior KD techniques for QAT to boost the accuracy of large Transformer encoders.
\item We quantitatively reveal that the \textit{attention-map loss} (based on KL-Div) outperforms the existing attention-score loss (based on MSE). The proposed attention-map loss is particularly beneficial for the BERT-Base model.
\item We discover the task-dependent attention characteristics, particularly noticeable in BERT-Large. In particular, we reveal that specific tasks on large Transformers suffer homogenization of attention output when weights are quantized. We propose a new KD method, \textit{attention-output loss}, to address this issue.
\item We further explore the potential of \textit{unifying the attention-map and output losses} to handle task-dependent attention characteristics ubiquitously.
\item We evaluate the proposed KD methods on various large-scale Transformer encoders and NLP tasks, achieving state-of-the-art accuracy for sub-2-bit aggressive QAT. 
\end{itemize}

%\subfile{Sections/intro.tex}
% --------------------------------------------------------------------------- %
% Related Work and discussion
% --------------------------------------------------------------------------- %

\begin{figure}[H]
\centerline{\includegraphics[width=0.8\columnwidth]{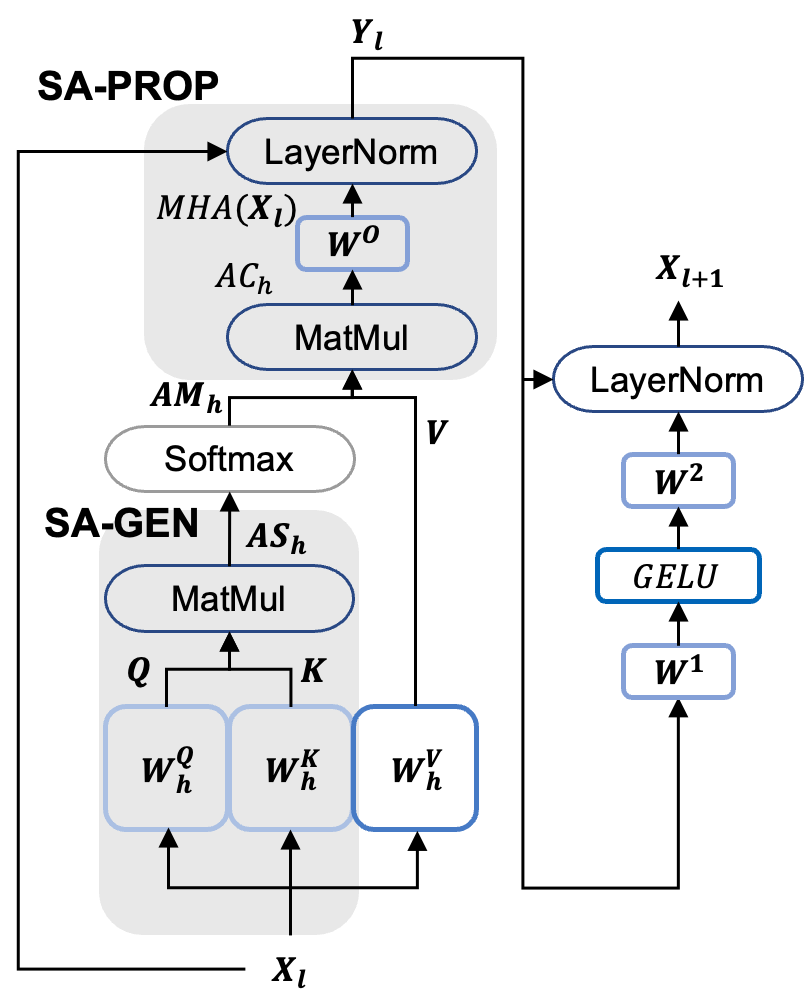}
}
\caption{The Transformer layer architecture.}
\label{fig:attention_layer}
\end{figure}

\section{Related Work}
\label{sec:related}

\subsection{Transformer Encoder Model}
\label{subsec:background_transformer}

Transformer-based encoder models like BERT~\cite{devlin2018bert} has been widely adopted for natural language processing (NLP) tasks such as question answering and language inference. As Fig.~\ref{fig:attention_layer} shows, these models are built with Transformer layers consisting of Multi-Head Attention (MHA) and Feed-Forward Network (FFN)~\cite{vaswani2017attention}. The input to the $l$-th Transformer layer is $\textbf{X}_l\in \mathbb{R}^{n\times d}$ where $n$ and $d$ are the sequence length and hidden state size, respectively. Let $N_H$ be the number of attention heads and $d_h=d/N_H$. $\textbf{W}^{Q}_h,\textbf{W}^{K}_h,\textbf{W}^{V}_h\in \mathbb{R}^{d\times d_h}$ are the weight parameters converting $\textbf{X}_l$ into Query ($\textbf{Q}=\textbf{X}_l \textbf{W}^{Q}_h$), Key ($\textbf{K}=\textbf{X}_l \textbf{W}^{K}_h$), and Value ($\textbf{V}=\textbf{X}_l \textbf{W}^{\textbf{V}}_h$), respectively. Then, attention score ($\text{\textbf{AS}}_h = \textbf{Q}\textbf{K}^{\top}$), attention map ($\text{\textbf{AM}}_h = \text{Softmax}_h(\frac{\text{\textbf{AS}}_h}{\sqrt{d}})$), and attention context ($\text{\textbf{AC}}_h = \text{\textbf{AM}}_h\textbf{V}$). 

MHA is defined as:
\begin{equation}
\begin{aligned}
\text{MHA}(\textbf{X}_l)=\text{Concat}(\text{\textbf{AC}}_1,...\text{\textbf{AC}}_{N_H})\textbf{W}^O.
\end{aligned}
\end{equation}
Motivated by \cite{not_only_a_weight}, MHA can be re-written per each token $i$:
\begin{equation}
\begin{aligned}
\text{MHA}(\textbf{X}_l)(i) = \sum_{j=1}^{n} \alpha_{i,j}f(\textbf{X}_{l}(j)),
\end{aligned}
\end{equation}
where $f(x) := (x\textbf{W}^{V} + \textbf{b}^{V})\textbf{W}^{O}$ and $\alpha_{i,j}$ is $j$'th attention probability of $i$'th token in \textbf{AM}$_h$. Therefore, MHA can be decomposed into two parts: self-attention generation (SA-GEN) corresponding to the attention map ($\alpha$), and self-attention propagation (SA-PROP) corresponding to $f(x)$. Fig.~\ref{fig:attention_layer} shows which part is SA-GEN and SA-PROP respectively.

FFN consists of two fully-connected layers with weight parameters $\textbf{W}^1$ and $\textbf{W}^2$: 
\begin{equation}
\begin{aligned}
\text{FFN}(\textbf{Y}_l)=\text{GELU}(\textbf{Y}_{l}\textbf{W}^1+\textbf{b}^1)\textbf{W}^2+\textbf{b}^2.
\end{aligned}
\end{equation}
Therefore, output of a Transformer layer $\textbf{X}_{l+1}$ is defined as:
\begin{equation}
\begin{aligned}
\label{eq:layernorm}
\textbf{Y}_l&=\text{LayerNorm}(\textbf{X}_l+\text{MHA}(\textbf{X}_l)), \\
\textbf{X}_{l+1}&=\text{LayerNorm}(\textbf{Y}_l+\text{FFN}(\textbf{Y}_l)).
\end{aligned}
\end{equation}
Here, $\textbf{Y}_l$ and $\textbf{X}_{l+1}$ are called attention output (\textbf{AO}) and Transformer output, respectively.

% --------------------------------------------------------------------------- %
% Knowledge Distillation for BERT Compression
% --------------------------------------------------------------------------- %

\subsection{Knowledge Distillation for Compression of Transformer Models}

Knowledge distillation (KD)~\cite{kd} is a transfer learning framework that a lightweight model (the student) learns from the knowledge distilled from a cumbersome model (the teacher). Since KD provides the student information to reach the teacher's capability, KD has been widely adopted for model compression of large-scale Transformer models like BERT. A basic distillation approach is to match the probability distribution at the output of the teacher and student models via CE loss, as in DistilBERT~\cite{distilbert}. In addition to this soft-label distillation, PKD~\cite{pkd} suggested KD on the normalized output of each Transformer layer, as distillation on the teacher's intermediate representations can benefit the student. MobileBERT~\cite{mobilebert} also employed per-head KD on the attention map and customized architecture for efficient Transformer computations. MiniLM and MiniLMv2~\cite{minilm,wang2021minilmv2} further transferred relational knowledge from the self-attention map, but only at a single Transformer layer (located at the last or upper-middle). \cite{marginal} further claimed that distilling more intermediate representations does not necessarily help improve the accuracy of the student. 

Although these KD-based compression techniques have developed efficient BERT structures, there has been limited understanding of KD on the model quantization. In particular, we are the first to quantitatively reveal that more distillation of the intermediate representations helps QAT reduce the accuracy gap between the quantized student and the full-precision teacher.

% --------------------------------------------------------------------------- %
% Quantization for BERT
% --------------------------------------------------------------------------- %
\subsection{Quantization for BERT}
Quantization is a promising technique to reduce the high inference cost of large-scale models without changing the model structure.
Instead of representing numbers in 32-bit floating-point (FP32), employing fixed-point representation, such as 8-bit integer (INT8) quantization, has achieved significant speedup and storage savings for BERT~\cite{q8bert, ibert, lin2021fq}. However, direct quantization of weight parameters leads to degradation of the original model accuracy when quantization bit-precision is low. Therefore, quantization-aware training (QAT) is commonly applied for ultra-low precision model quantization. 

Recently, QAT has been applied for compressing BERT with a precision lower than 2-bit. TernaryBERT~\cite{ternarybert} represents each weight parameter into one of three values $\{-1, 0, 1\}$. TernaryBERT actively incorporates KD into QAT for improving accuracy degradation. To reduce the bit-precision, BinaryBERT~\cite{binarybert} proposed a modified QAT procedure with a specific weight initialization for binary quantization. DQ-BART~\cite{li2022dq} further combined model compression (via layer reduction) and quantization by exploiting KD. 

Although KD has been a de-facto technique for QAT, there is a lack of understanding about why. In particular, the aforementioned QAT methods all employed the layer-wise KD on the self-attention score ($\text{AS}_l$) and Transformer output ($X_l$) along with the KD on soft labels. Considering numerous KD techniques with various choices for the knowledge sources and the objective, it is not clear if the current recipe helps QAT the most. This work investigates the prior layer-wise KD techniques and improves them with new objectives and knowledge sources.

% % % --------------------------------------------------------------------------- %
% % % Background on Transformer and KD with QAT 
% % % --------------------------------------------------------------------------- %
% \input{Sections/background.tex}

% % --------------------------------------------------------------------------- %
% % Exploration in KD
% % --------------------------------------------------------------------------- %
\section{Prior KD Techniques for QAT}
\label{sec:exploration}

In this section, we investigate prior KD techniques for QAT evaluated on BERT-Base. As discussed earlier, KD techniques commonly used for QAT include 1) all-layer distillation and 2) distillation on SA-GEN. First, we provide justification and improvement on these techniques. Then we further showcase the limitation when they are applied to large-scale Transformer encoders.

\subsection{All-Layer Distillation for QAT}
\label{subsec:l2l}

\begin{figure}[t]
\begin{center}

\centerline{\includegraphics[width=1\columnwidth]{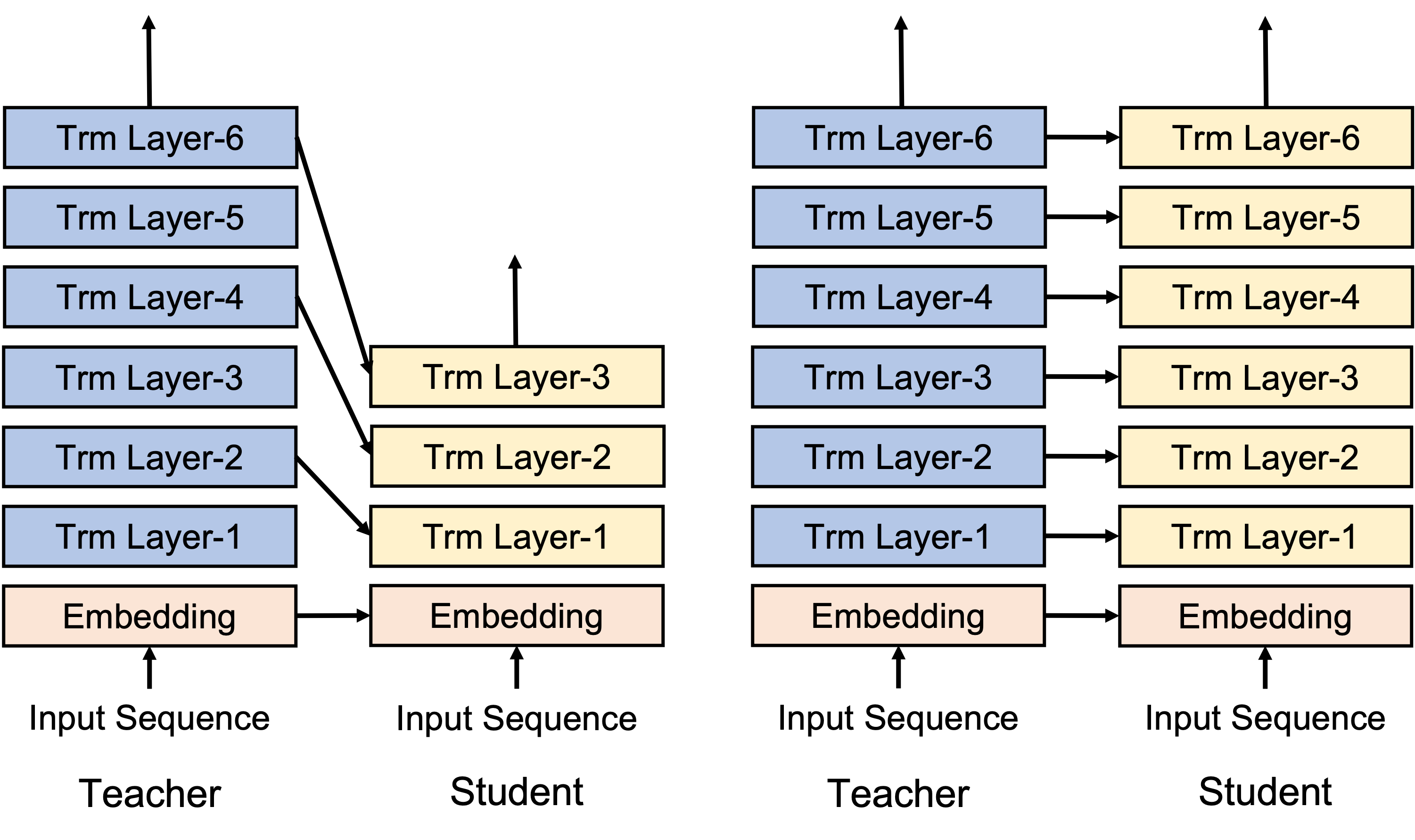}}
\caption{Illustration of layer selection strategy in model compression and model quantization. Left: Uniform mapping strategy. Right: All-layer distillation.}
\label{fig:strategy}
\end{center}
\vskip-0.3in
\end{figure}

Generally, the internal representation of the teacher, such as a layer output, is widely used for knowledge distillation for model compression~\cite{aguilar2020knowledge}. However, there is a distinct difference in KD between typical model compression and quantization. For example, Fig.~\ref{fig:strategy} shows two representative layer-to-layer mapping for KD: selected-layer distillation for model compression (left) and all-layer distillation for QAT (right). In the case of selected-layer distillation, the study showed that the marginal utilities of hidden state knowledge (HSK) diminish as more HSK has been distilled\cite{marginal}. In contrast, most prior QAT methods applied KD on the Transformer output of all the layers. The structural equivalence of the teacher and the student of QAT methods makes this choice natural, but there is little justification. 

\begin{figure}[t]
\begin{center}
\centerline{\includegraphics[width=1\columnwidth]{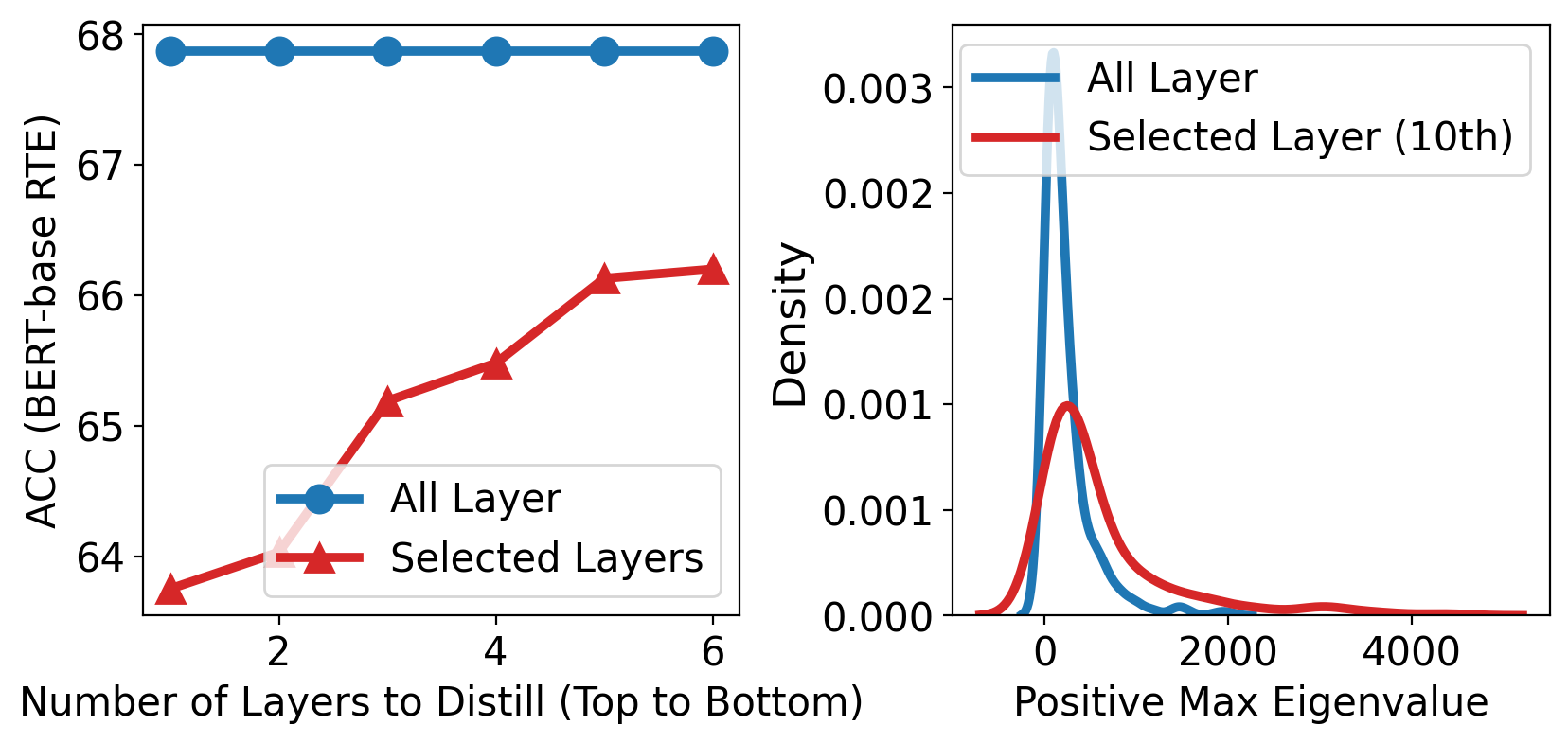}}
\caption{Comparison of (a) RTE accuracy and (b) Hessian max Eigenvalues Spectra between the selected-layer distillation and all-layer distillation in BERT-Base.}
\label{fig:grouping}
\end{center}
\end{figure}

We conjecture that quantization applied to the weight parameters disrupts the functionality of the Transformer layer, necessitating layer-wise guidance. To validate this conjecture, we conducted two experiments. First, we compared the accuracy of the uniformly selected layer distillation with a varying number of distilled layers. As shown in Fig.~\ref{fig:grouping}a, the accuracy grows along with the number of distilled layers, and all-layer distillation significantly outperforms selected-layer distillation. In addition, we compared the loss surface of the two distillation approaches after QAT in terms of Hessian max eigenvalues~\cite{park2021vision}. We adopt single layer selection distillation as a strategy of selected layer distillation. In particular, we used the method of selecting the 10th layer of BERT-Base model, which was most helpful in performance as a single layer selection distillation. As shown in Fig.~\ref{fig:grouping}b, all-layer distillation shows smaller magnitudes of Eigenvalues, indicating a smoother loss surface. Therefore, we can conclude that the layer-wise distillation helps train the student with the quantized weight parameters.

\subsection{Improve KD on Self-Attention Generation}
\label{subsec:KDonSA}

We further investigate the objective of all-layer KD. As discussed earlier, prior QAT methods employed MSE loss on the attention score (called attention-score loss) for all-layer KD, as follows:
\begin{equation}
\begin{aligned}
\label{eq:mse_attn_loss}
\mathcal{L}_{score} = \sum_{l=0}^{L-1} \text{MSE}(\text{\textbf{AS}}_{l}^{T}, \text{\textbf{AS}}_{l}^{S}).
\end{aligned}
\end{equation}
Given that the attention map captures the correlation of one token to all the others, it is essential to maintain the relative importance of tokens. However, quantization in nature clamps and coarsely represents the weight parameters, making attention less distinguishable. We expect KD to help maintain disparity, but the attention-score loss is not a proper objective since it mainly focuses on logit matching~\cite{kl_and_mse}. 

As an alternative, we propose to use the KL-Div loss on the attention-map (called attention-map loss) defined as follows:
\begin{equation}
\begin{aligned}
\label{eq:kl_attn_loss}
\mathcal{L}_{map} = \frac{1}{N_{h}n}\sum_{h=1}^{N_h} \sum_{t=1}^{n}
D_{KL}(\text{\textbf{AM}}_{l,h,t}^{T}\|\text{\textbf{AM}}_{l,h,t}^{S}).
\end{aligned}
\end{equation}
Assuming that the temperature hyper-parameter ($\tau$) is one, KL-Div focuses on label-matching \cite{kl_and_mse}. Thus the relative importance of attention across tokens is better maintained with attention-map loss. Although the attention-map loss was previously employed in model compression~\cite{wang2021minilmv2}, we are the first to quantitatively reveal the benefits of the attention-map loss in the context of QAT.

\begin{figure}[t]
% \begin{wrapfigure}{r}{6cm}
\begin{center}
\centerline{\includegraphics[width=1\columnwidth]{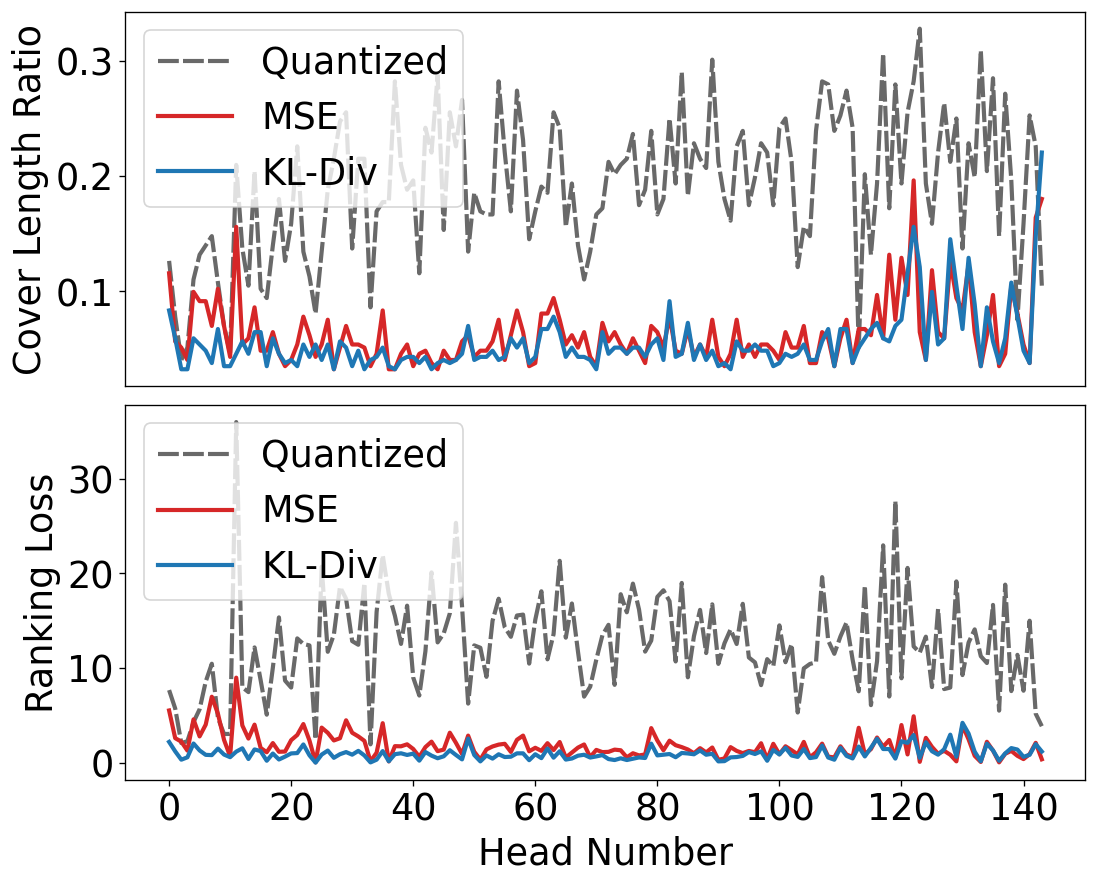}}
\caption{The cover length ratio and ranking loss per attention head in BERT-Base. X-axis: Number of attention heads. Quantized: Quantized model without applying QAT, KL-Div: attention-map loss, MSE: attention-score loss}
\label{fig:cover_and_ranking}
\end{center}
\vskip-0.2in
\end{figure}

We introduced two metrics that characterize the attention map to evaluate the proposed KD loss quantitatively. The cover length ratio captures the student's attention map deviation from the teacher's based on Top-K token coverage. The ranking loss~\cite{ptq4vit} shows the similarity in the attention rankings of the teacher and the student. Fig.~\ref{fig:cover_and_ranking} compares the cover length ratio and the ranking loss of 
every attention head of BERT-Base on SST-2 task (the overall trend is the same for the other layers and tasks). As shown in the figure, quantization significantly increases the cover length ratio and the ranking loss, indicating that the relative ratings of the attention are seriously distorted. The attention-score loss helps reduce such distortion, yet spikes still exist. In contrast, attention-map loss successfully suppresses the spikes, maintaining the relative importance of the attention map. More details about cover length ratio and ranking loss are described in Appendix.~\ref{appen:cover_ranking}.

To further understand the impact of the objectives of KD on the QAT accuracy, we conducted the temperature sweep of KL-Div. Since the gradients of KL-Div loss can be simplified into the gradients of MSE loss when the temperature is sufficiently large ~\citep{kl_and_mse}, we can manage the behavior of KL-Div loss through sweeping the temperature value ($\tau$), where $\tau$=1 and $\tau=\inf$ resemble the attention-map and attention-score losses, respectively. Table~\ref{tab:kldiv_temp} shows the QAT accuracy of BERT-Base on CoLA and STS-B with varying $\tau$. As shown in the table, the accuracy of the quantized model increases as the loss term becomes similar to the attention-map loss. Such performance improvement supports our understanding that 1) label matching is crucial for compensating QAT on SA-GEN, and 2) the attention-map loss is more effective for label matching.

\begin{table}[H]
\centering
\resizebox{1\linewidth}{!}{
\begin{tabular}{lcccccc}
\toprule
      &                 \multicolumn{6}{c}{KL Divergence Temperature Hyper-Parameter ($\tau$)}           \\
 
Task  & KL Div         & $\tau$=1       & $\tau$=5      & $\tau$=10      & $\tau$=20      & MSE \\
\midrule
CoLA  & \textbf{50.76} & 50.76 & 49.11         & 47.51 & 47.19          & 47.51          \\
STS-B & \textbf{87.78} & 87.69 & 87.20 & 87.19          & 87.29 & 87.55 \\
\bottomrule
\end{tabular}}
\caption{Attention based KD-QAT with KL Div Temperature Hyper Parameter sweeping on CoLA and STS-B tasks with BERT-Base (Each experiment is repeated 5 times.)}
\label{tab:kldiv_temp}
\end{table}

% % --------------------------------------------------------------------------- %
% % Methods 
% % --------------------------------------------------------------------------- %

\section{KD for QAT on Large Transformers}
\label{sec:KDonLarge}

We extend the investigation of KD techniques to QAT on large transformer models. In this section, we first reveal the limitation of the attention-map loss due to the task-dependent characteristics. Then we propose a new KD loss, the attention-output loss, to address this challenge. Lastly, we propose a combination of the two losses to handle task-dependent characteristics. 

\subsection{Task-Dependent Characteristics}
\label{subsec:task-dependent}

Although the same pre-trained models are employed for the downstream fine-tuning, the characteristics of attention vary depending on the tasks ~\cite{kovaleva-etal-2019-revealing}. Motivated by the discussions of \cite{bondarenko-etal-2021-understanding} that outliers in activations of residual connections (i.e., SA-PROP) arrange specific attention patterns, we examine these outliers via min-max curves at attention output to understand task-dependent characteristics for quantization. 

\begin{figure}[t]
\leftline{\includegraphics[width=1\columnwidth]{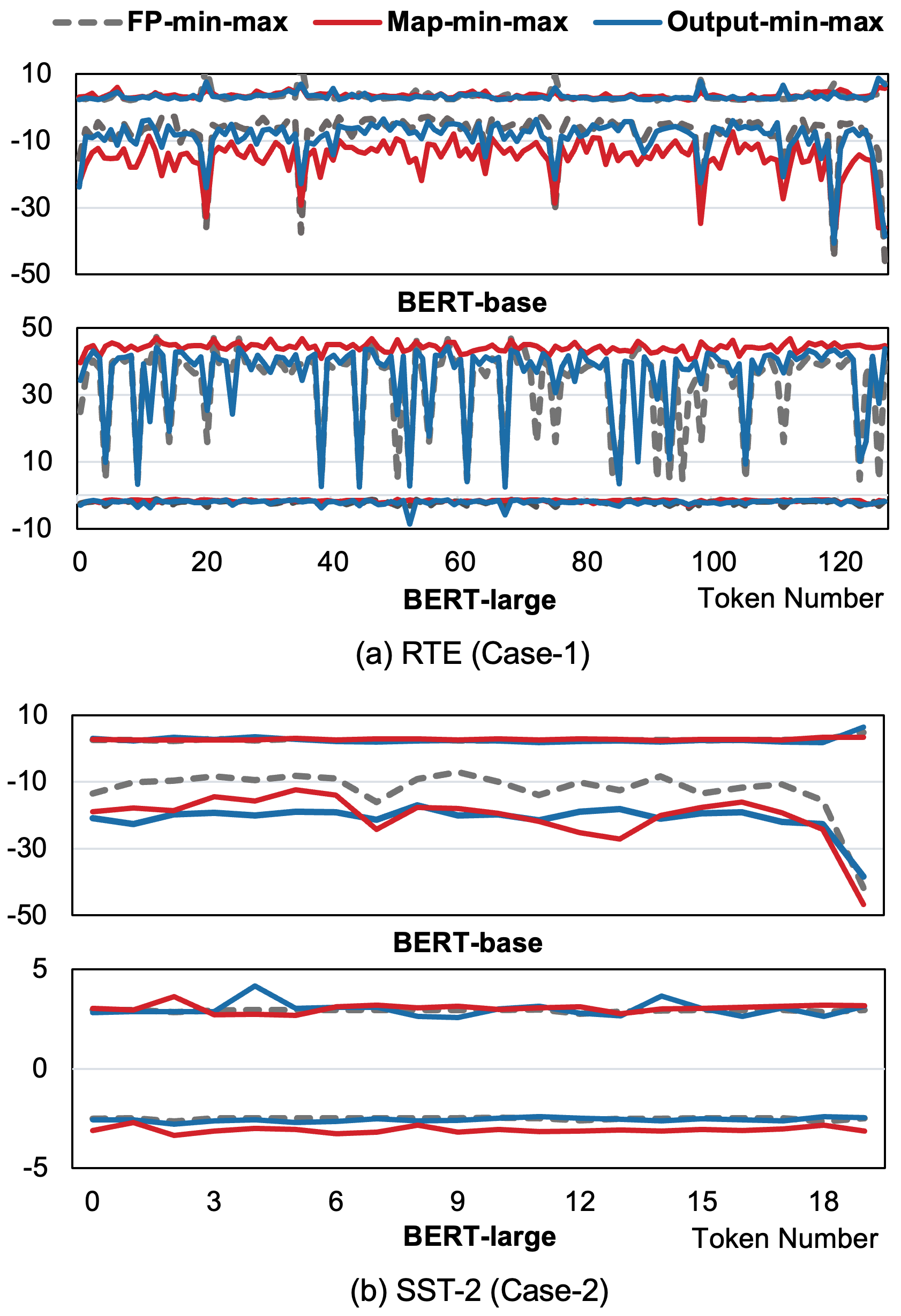}}
\caption{Comparison of per-token dynamic ranges of attention output ($Y_l$) between BERT-Base (top) and BERT-Large (bottom) for RTE and SST-2. Each pair of curves delineates min-max values at the token's attention output. FP-min-max and Map/Output-min-max correspond to the min-max curves of the teacher model and the student with the attention map/output loss, respectively.}
\label{fig:per_token_min-max}
\vskip-0.2in
\end{figure}

Fig.~\ref{fig:per_token_min-max} plots the dynamic range of the attention output ($Y_l$) across the input tokens. There are two representative cases: Case-1) the task with distinct attention values (especially for special tokens) such as RTE, and Case-2) the task with homogeneous attention values such as SST-2. Each case's overall attention characteristics are intensified as the model size increases. For example, distinct features of RTE's attention become more drastic on BERT-Large. 

Since the quantization clamps and coarsely represents the values, it is challenging to maintain the distinct attention for the tasks in Case-1. As discussed in Sec.~\ref{subsec:KDonSA}, in the case of BERT-Base, the attention-map loss was capable of recovering the disparity in the attention (Fig.~\ref{fig:per_token_min-max}a-Top). However, as shown in Fig.~\ref{fig:per_token_min-max}a-Below, the attention-map loss failed to adjust the attention in the case of BERT-Large.

We conjecture that the attention-map loss fails due to the increased number of layers of BERT-Large. We adopt the analysis framework of \cite{not_only_a_weight} to separately analyze the layer-wise behavior of SA-GEN and SA-PROP. Fig.~\ref{fig:tr_norm} plots the average distance of SA-GEN and SA-PROP of a special token~\textit{[SEP]} from the teacher for RTE (Case-1) and SST-2 (Case-2) with BERT-Large. Note that the attention-map loss suppresses the distance in SA-GEN. This suppression of SA-GEN deteriorates the attention output (c.f., the attention-map loss is effective for SST-2). With many layers, quantization along SA-PROP fails KD with attention-map loss to recover distinctive attention. Therefore, we need a new KD loss to handle disruption from SA-PROP.

\subsection{Attention Output Loss}
\label{subsec:output_loss}

\begin{figure}[t]
\begin{centering}
\centerline{\includegraphics[width=1\columnwidth]{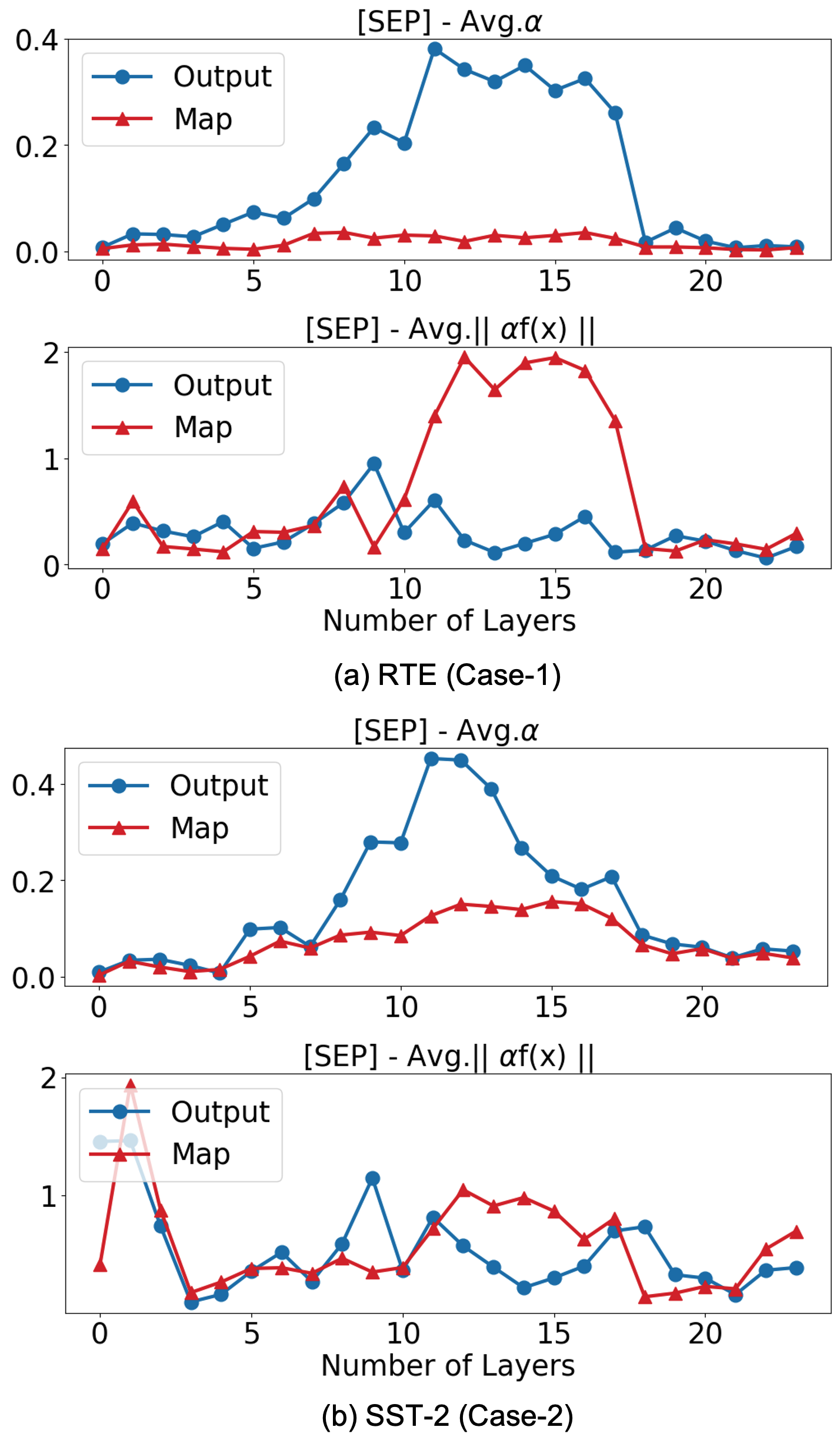}}
\caption{Average distance of SA-GEN(self-attention probability - $\alpha$) and SA-PROP(self-attention propagation -$f(x)$) from the teacher model in two tasks (RTE, SST-2) with BERT-Large.}
\label{fig:tr_norm}
\end{centering}
\vskip-0.2in
\end{figure}

Observations from Fig.~\ref{fig:tr_norm} imply that SA-PROP becomes a disruption source for QAT of BERT-Large on Case-1 tasks. One way to suppress the quantization error along SA-PROP is to apply KD directly to SA-PROP. Therefore, we devise a new KD loss, the attention-output loss as follows:
\begin{equation}
\begin{aligned}
\label{eq:attn_output_loss}
\mathcal{L}_{output} = \sum_{l=0}^{L-1} \text{MSE}(\textbf{Y}_{l}^{T}, \textbf{Y}_{l}^{S}).
\end{aligned}
\end{equation}

The benefits of the attention-output loss are apparent. As shown in Fig.~\ref{fig:per_token_min-max}a, the attention output with the attention-output loss follows the distinctive attention of the full-precision teacher. We can understand the mechanism of the attention-output loss via Fig.~\ref{fig:tr_norm}; the attention-output loss allows modification of SA-GEN to adjust the attention map so that the resulting attention output matches better with the teacher. Note that the change in SA-GEN occurs at the upper layers of the Transformer models; thus, the attention-output loss is more beneficial for large Transformer models. 

\begin{figure}[t]
\begin{center}
\centerline{\includegraphics[width=1\columnwidth]{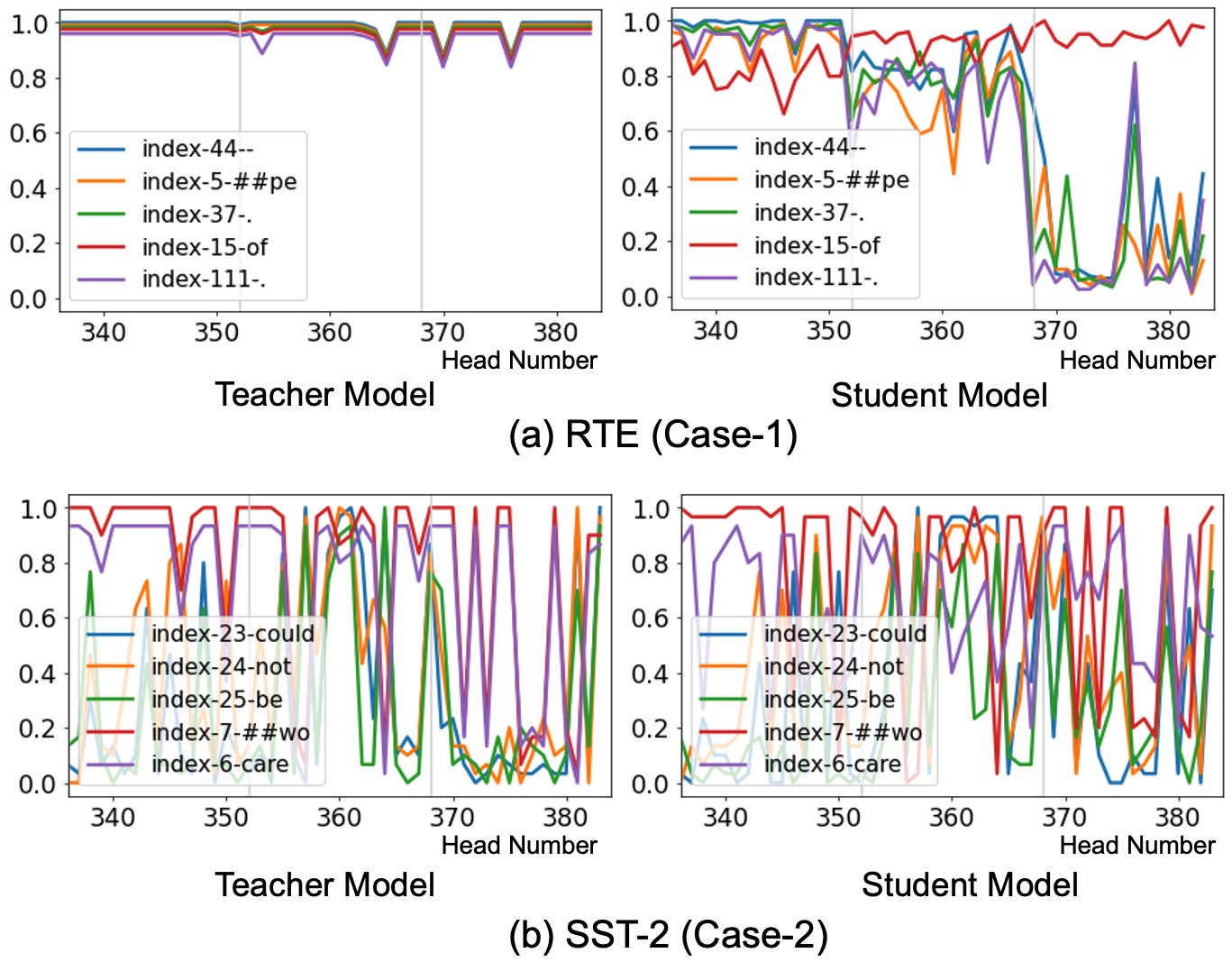}}
\caption{Ranking ratio per head (Last three layer's 48 heads) on the SST-2 and RTE task with BERT-Large. Left: Teacher Model, Right: Student Model}
\label{fig:ranking_ratio}
\end{center}
\vskip-0.3in
\end{figure}

To further understand the task-dependent characteristics, we empirically observe the attention-output loss's impact on the attention map's self-attention probability. To quantify the modification in the attention map, we introduce the ranking ratio, defined as a ranking of the attention probability of an individual token normalized by the sequence length. Fig.~\ref{fig:ranking_ratio} tracks the ranking ratio of the selected tokens of full-precision teacher and quantized student models per each head. In the case of RTE (i.e., Case-1 task), QAT with the attention-output loss exhibits rapid changes in ranking in a specific direction toward reduced attention-output loss. In the case of SST-2, however, the situation is very different; rankings of the selected tokens change significantly regardless of KD for QAT. Thus, KD on the attention output cannot drive the rankings in any meaningful direction. These observations confirm the importance of considering task-dependent characteristics for successful KD for QAT. 

\subsection{Unified Attention-Map and Output Loss}
\label{subsec:scaled-mix}
Considering task-dependent attention characteristics of BERT-Large, we further explore the potential of unifying the attention-map and output losses for QAT. Note that the preference between the attention-map and output losses varies according to the model size (e.g., BERT-Base vs. Large) and tasks (Case-1 vs. Case-2). As for exploration, we formulate a unified attention-map and output loss with $\gamma$ as a mixing parameter as follows:
\begin{equation}
\begin{aligned}
\label{eq:mixed_scale}
\mathcal{L}_{SM_{1}} &= \mathcal{L}_{map} + ~\gamma \mathcal{L}_{output}, \\
\mathcal{L}_{SM_{2}} &= \mathcal{L}_{output} + ~\gamma \mathcal{L}_{map}, \\
\text{where}\ &\gamma ~\in \{0.1, 0.2, 0.3, \dots,0.9\}.
\end{aligned}
\end{equation}

As will be discussed in Sec.\ref{subsec:bert_expr}, the unified loss can boost the accuracy of the best performing KD loss (either the attention-map or output loss). As applying this unified loss in KD-QAT, we identified that every tasks has its own score favorable mixing parameters which shows task-dependent characteristics. Detailed mixing parameter information for each task is in Appendix.~\ref{appen:scale-mix_exploring}.

% % --------------------------------------------------------------------------- %
% % Experiments & Ablation Study
% % --------------------------------------------------------------------------- %
\section{Experiments}
\label{sec:expr}

\subsection{Experimental Setup}
\label{subsec:expr_setting}

\begin{table*}[t]
\centering
\resizebox{1\linewidth}{!}{

\begin{tabular}{clllllllll}
\toprule
  GLUE Task & RTE$^{\dagger}$             & CoLA$^{\dagger}$     & STS-B$^{\dagger}$ & SST-2$^{\star}$           & QNLI$^{\star}$           & MNLI$^{\star}$           & QQP$^{\star}$      & MRPC         & AVG   \\

(Dataset)       & (2.5k)          & (8.5k)         & (5.7k)         & (67k)         & (108k)           & (393k)         & (364k)         & (3.5k)         &       \\
\midrule
Full-Prec    & 73.28           & 58.04          & 89.24           & 92.09           & 91.32          & 84.37          & 89.30        & 87.77  & 83.39 \\
\midrule    
Baseline     & 68.53 {\color{gray}\scriptsize±1.69}           & 49.61 {\color{gray}\scriptsize±0.79}         &87.55 {\color{gray}\scriptsize±0.14}                   & 92.01 {\color{gray}\scriptsize±0.29}            &90.65 {\color{gray}\scriptsize±0.05}          &84.21 {\color{gray}\scriptsize±0.10}          & 89.06 {\color{gray}\scriptsize±0.40}        & \textbf{88.58} {\color{gray}\scriptsize±0.40} & 81.28  \\
\midrule
Map          & 70.39 {\color{gray}\scriptsize±0.78}           & 50.40 {\color{gray}\scriptsize±1.03}         & \textbf{87.78} {\color{gray}\scriptsize±0.15}         & 92.13 {\color{gray}\scriptsize±0.22}         & \textbf{90.98} {\color{gray}\scriptsize±0.17}          &84.31 {\color{gray}\scriptsize±0.10}          & 89.22 {\color{gray}\scriptsize±0.40}        & 88.07 {\color{gray}\scriptsize±0.40} & 81.66  \\
Output     & 70.65 {\color{gray}\scriptsize±1.27}           & 49.05 {\color{gray}\scriptsize±0.50}           & 87.77 {\color{gray}\scriptsize±0.14}           & 92.13 {\color{gray}\scriptsize±0.22}           & 90.58 {\color{gray}\scriptsize±0.07}           & 84.24 {\color{gray}\scriptsize±0.01}          & 89.17 {\color{gray}\scriptsize±0.20}        & 87.01 {\color{gray}\scriptsize±0.43} & 81.33  \\

Map+Output & \textbf{71.68} {\color{gray}\scriptsize±1.19}           & \textbf{50.50} {\color{gray}\scriptsize±0.45}         &87.73 {\color{gray}\scriptsize±0.16}                   & \textbf{92.39} {\color{gray}\scriptsize±0.18}            &90.91 {\color{gray}\scriptsize±0.14}          &\textbf{84.33} {\color{gray}\scriptsize±0.06}          & \textbf{89.28} {\color{gray}\scriptsize±0.10}        & 88.18 {\color{gray}\scriptsize±0.53} & \textbf{81.87}  \\

\bottomrule
\end{tabular}}

\caption{\label{bert-base_table}
BERT-Base: Performance of KD-QAT Results on GLUE benchmark (8-bit activation and ternary weight quantization, the compression rate of quantized BERT-Base is 14.9x). Small dataset (under 10k) tasks are repeated 5 times; the others are repeated 3 times. $\dagger$ and $\star$ indicate Case-1 and Case-2 GLUE tasks respectively.
}
\end{table*}

\begin{table*}[t]
\centering
\resizebox{1\linewidth}{!}{

\begin{tabular}{clllllllll}
\toprule
   GLUE Task & RTE$^{\dagger}$             & CoLA$^{\dagger}$     & STS-B$^{\dagger}$ & SST-2$^{\star}$           & QNLI$^{\star}$           & MNLI$^{\star}$           & QQP$^{\star}$      & MRPC         & AVG   \\

(Dataset)       & (2.5k)          & (8.5k)         & (5.7k)         & (67k)         & (108k)           & (393k)         & (364k)         & (3.5k)         &       \\
\midrule
Full-Prec   & 70.39  & 60.31  & 89.83  & 92.32  & 92.29  & 86.49  & 89.55  & 88.43  & 83.70 \\
\midrule    
Baseline     & 65.02 {\color{gray}\scriptsize±1.40}           & 52.87 {\color{gray}\scriptsize±0.99}         &88.75 {\color{gray}\scriptsize±0.09}        & 91.82 {\color{gray}\scriptsize±0.22}        &91.87 {\color{gray}\scriptsize±0.15}        & 85.70 {\color{gray}\scriptsize±0.17}          & 89.29 {\color{gray}\scriptsize±0.07}        & \textbf{89.26} {\color{gray}\scriptsize±0.54} & 81.84  \\
\midrule
Map          & 66.42 {\color{gray}\scriptsize±0.75}           & 53.16 {\color{gray}\scriptsize±0.53}         & 88.65 {\color{gray}\scriptsize±0.11}         & 92.20 {\color{gray}\scriptsize±0.30}         & 91.93 {\color{gray}\scriptsize±0.13}          &86.10 {\color{gray}\scriptsize±0.13}          & \textbf{89.53} {\color{gray}\scriptsize±0.07}        & 88.67 {\color{gray}\scriptsize±0.37} & 81.28  \\
Output     & \textbf{69.50} {\color{gray}\scriptsize±1.20}           & \textbf{54.71} {\color{gray}\scriptsize±0.71}         & \textbf{89.10} {\color{gray}\scriptsize±0.08}         &92.13 {\color{gray}\scriptsize±0.26}         &91.92 {\color{gray}\scriptsize±0.13}          &86.22 {\color{gray}\scriptsize±0.05}          & 89.44 {\color{gray}\scriptsize±0.09}        & 88.75 {\color{gray}\scriptsize±0.71} & \textbf{82.72}  \\
Map+Output & 68.83 {\color{gray}\scriptsize±1.45}           & 54.69 {\color{gray}\scriptsize±1.08}         &88.85 {\color{gray}\scriptsize±0.15}        & \textbf{92.30} {\color{gray}\scriptsize±0.11}       & \textbf{92.16} {\color{gray}\scriptsize±0.15}          &\textbf{86.36} {\color{gray}\scriptsize±0.06}          & 89.48 {\color{gray}\scriptsize±0.06}        & 88.64 {\color{gray}\scriptsize±0.79} & 82.66  \\

\bottomrule
\end{tabular}}

\caption{\label{tab:bert-large_table}
BERT-Large: Performance of KD-QAT Results on GLUE benchmark (8-bit activation and ternary weight quantization, the compression rate of quantized BERT-Large is 15.4x). Small dataset (under 10k) tasks are repeated 5 times; the others are repeated 3 times. $\dagger$ and $\star$ indicate Case-1 and Case-2 GLUE tasks respectively.
}
\end{table*}

We employ three Transformer encoder models (BERT-Base, BERT-Large, ULM-Encoder-Large) to evaluate the proposed KD methods. BERT~\cite{devlin2018bert} consists of Transformer encoder layer, finetuned for GLUE downstream tasks~\cite{devlin2018bert}. ULM-Encoder-Large~\cite{kt-ulm} is a Korean language model based on T5~\cite{t5}, finetuned for KLUE downstream tasks~\cite{park2021klue}. 

The configuration of each model is as follows:
\begin{enumerate}
    \item \textbf{BERT-Base.} It is a 12-layer Transformer encoder with a hidden dimension of 768 using 12 attention heads and contains about 110M parameters.
    \item \textbf{BERT-Large.} It is composed of 24 Transformer encoder layers, and uses a hidden dimension of 1024 with 16 attention heads. This model contains about 340M parameters.
    \item \textbf{ULM-Encoder-Large.} It also has the same configuration as BERT-large except for feed-forward dimension, which is 2816 for ULM-Encoder-Large while BERT-Large has 4096. It contains about 280M parameters.
\end{enumerate}

We initiate QAT from the task-specific finetuned models. Our experiments were performed on A6000 GPUs. Our implementation is based on the TernaryBERT PyTorch codebase.\footnote{\url{ https://github.com/huawei-noah/Pretrained-Language-Model/tree/master/TernaryBERT}} All embedding and weight parameters are ternarized and the activations are quantized to 8-bit for QAT. we use layer-wise ternarization for weights in Transformer layers while row-wise for the word embedding, same as TernaryBERT ~\cite{ternarybert}. Also, all the experiments are repeated 5 times, unless stated otherwise. 

For performance comparison, we consider the following KD options:
\begin{itemize}
    \item \textbf{Baseline.} The standard TernaryBERT with the attention-score and Transformer output loss along with KD on soft labels.
    \item \textbf{Map.} Use the attention-map loss instead of the attention-score loss of TernaryBERT.
    \item \textbf{Output.} Use the attention-output loss instead of the attention-score loss of TernaryBERT.
    \item \textbf{Map+Output.} Use the unified attention-map and output loss instead of the attention-score loss of TernaryBERT.
\end{itemize}

\subsection{Experiments on BERT-Base and Large}
\label{subsec:bert_expr}

Tables~\ref{bert-base_table} and~\ref{tab:bert-large_table} show the result on the development set across the GLUE benchmark. Notable observations are summarized as follows:
\begin{itemize}
    \item The GLUE tasks can be categorized into two cases. Case-1($\dagger$): RTE, CoLA, STS-B. Case-2($\star$): SST-2, QNLI, MNLI, QQP. 
    \item In the case of BERT-Base, attention-map loss benefits all the tasks in Case-1 and Case-2, whereas attention-output loss is ineffective.
    \item In the case of BERT-Large, the attention-map loss is marginally helpful for Case-1 and Case-2, while the attention-output loss significantly boosts the accuracy of Case-1 tasks.
    \item Overall, the unified loss facilitates QAT accuracy, except for BERT-Large on Case-1 tasks (in which the attention-output loss works the best). 
    \item MRPC is a corner case; the QAT accuracy often outperforms the Full-Precision accuracy, implying that quantization noise regularizes the model favorably for this task.   
\end{itemize}

\subsection{Experiments on ULM-Encoder-Large}
\label{subsec:ulm_expr}

\begin{table}[t]
\centering
\resizebox{1\linewidth}{!}{
\begin{tabular}{cllll}
\toprule
Task           & KLUE-TC  &  KLUE-STS  &  NSMC    & AVG       \\
(Dataset)      & \multicolumn{1}{l}{(45k)}    &   \multicolumn{1}{l}{(11k)}    &  \multicolumn{1}{l}{(150k)}  &       \\
\midrule
Full-Prec  & 85.76  & 92.11 & 91.87 & 89.91  \\
\midrule
Baseline & 85.56 {\color{gray}\scriptsize±0.08} & 91.04 {\color{gray}\scriptsize±0.10} & 91.13 {\color{gray}\scriptsize±0.04} & 89.24 \\
\midrule
Map & 85.41 {\color{gray}\scriptsize±0.10} & \textbf{91.44} {\color{gray}\scriptsize±0.23} & 91.24 {\color{gray}\scriptsize±0.10} & 89.36 \\
Output & \textbf{85.63} {\color{gray}\scriptsize±0.23} & 91.03 {\color{gray}\scriptsize±0.11} & 91.39 {\color{gray}\scriptsize±0.15} & 89.35
\\
Map + Output  & 85.57 {\color{gray}\scriptsize±0.21} & 91.11 {\color{gray}\scriptsize±0.14} & \textbf{91.65} {\color{gray}\scriptsize±0.12} & \textbf{89.44} \\
\bottomrule
\end{tabular}}
\caption{\label{tab:ulm-large}ULM-Large: Performance of KD-QAT Results on KLUE and NSMC dev dataset}

\end{table}

Table~\ref{tab:ulm-large} summarizes the average results three times of evaluating our KD methods for QAT on ULM-Encoder-Large. Overall, ULM-Encoder-Large is quite robust to quantization, but our KD methods surpass the baseline (= TernaryBERT). More specifically, the attention-map loss is more effective on the KLUE-STS task, while the output loss outperforms the map loss in KLUE-TC and NSMC, as shown in the table. Furthermore, the unified loss achieves the best accuracy on NSMC. Therefore, the proposed KD losses could improve the accuracy of the baseline QAT method.

\subsection{Ablation Study}
\label{subsec:ablation}

\begin{table}[t]
\centering
\resizebox{1\linewidth}{!}{
\begin{tabular}{cccccc}
\toprule
         GLUE Task & RTE$^{\dagger}$   & CoLA$^{\dagger}$  & STS-B$^{\dagger}$ & SST-2$^{\star}$ & QNLI$^{\star}$  \\
\midrule
Full-Prec             & 70.39 & 60.31 & 89.83 & 92.32 & 92.29 \\
\midrule
Map       & 66.42 & 53.16 & 88.65 & \textbf{92.20}  & \textbf{91.93} \\
MHA loss          & 66.78 & 54.01 & 88.69 & 92.08 & 91.84 \\
MHA loss + Residual & \textbf{69.50}  & \textbf{54.71} & \textbf{89.10}  & 92.13 & 91.92 \\
\bottomrule
\end{tabular}}

\caption{\label{tab:ablation-sa} Ablation study on the attention-output loss decomposing its sources into MHA({$\textbf{X}_{l}$}) and residual path ($\textbf{X}_{l}$) with BERT-Large on the GLUE tasks. \textdagger \, and $\star$ \, indicate Case-1 and Case-2 GLUE tasks respectively.}
\vskip-0.2in
\end{table}

In Sec.~\ref{subsec:output_loss}, we proposed attention-output loss to suppress the quantization error along the SA-PROP. As shown in Fig.~\ref{fig:attention_layer} and defined in Eq.~\ref{eq:layernorm}, attention-output loss integrates two sources of SA-PROP: \text{MHA}({$\textbf{X}_{l}$}) and residual connection($\textbf{X}_{l}$). We investigate attention-output loss's effectiveness by employing one of its two parts solely as a loss function objective: \text{MHA}({$\textbf{X}_{l}$}) (We call this loss function MHA loss). Specifically, MHA loss uses \text{MHA}($\textbf{X}_{l}$) as a objective of loss function instead of $\textbf{Y}_{l}$ in Eq.~\ref{eq:attn_output_loss}. \newline

 Table.~\ref{tab:ablation-sa} shows that the MHA loss method improves performance marginally in Case-1 tasks. When the residual connection is added to the MHA loss objective (MHA loss $+$ Residual in Table.~\ref{tab:ablation-sa}), which is equivalent to attention-output loss, the performance of all tasks increases. (especially in Case-1 GLUE tasks). These observations indicate that incorporating residual connection as an objective of attention-output loss is significant in recovering disruption of SA-PROP under the quantization.

% % --------------------------------------------------------------------------- %
% % Limitation & Conclusion
% % --------------------------------------------------------------------------- %

\section{Conclusion}
\label{sec:conclusion}

In this work, we investigate the mechanism of Knowledge distillation (KD) for QAT of large Transformers. We propose two KD methods, attention-map, and attention-output losses, to improve the recovery of the self-attention information. The experimental results on various Transformer encoder models demonstrate that the proposed KD methods and their combination achieve state-of-the-art accuracy for QAT with sub-2-bit weight quantization. Our code is available at \href{https://github.com/MarsJacobs/kd-qat-large-enc}{https://github.com/MarsJacobs/kd-qat-large-enc}.

\section{Limitation}
\label{sec:limitation}

This work investigates how KD works for QAT on Transformer Encoders. Although the analysis techniques employed in this work reveal many exciting insights, a more theoretical analysis of the impact of quantization under KD would be highly appreciated. Also, we explore the potential of unifying the two proposed KD techniques; incorporating automatic balancing of the two (or more) KD losses would be an interesting future research direction. 

\section*{Acknowledgement}
\label{sec:acknowledgement}

This work was partly supported by Institute of Information \& communications Technology Planning \& Evaluation (IITP) grants funded by the Korea government(MSIT) (No. 2020-0-01373, Artificial Intelligence Graduate School Program (Hanyang University), and No. 2022-0-00971, Logic Synthesis for NVM-based PIM Computing Architecture).

% Entries for the entire Anthology, followed by custom entries
\bibliography{anthology,custom}
\bibliographystyle{acl_natbib}

% % --------------------------------------------------------------------------- %
% % Appendix
% % --------------------------------------------------------------------------- %

\clearpage
\appendix

\section{Appendix}
\label{sec:appendix}

\subsection{Cover Length Ratio and Ranking Loss}
\label{appen:cover_ranking}

As shown in Sec.~\ref{subsec:KDonSA}, cover length ratio and ranking loss are metrics indicating how much the relative importance within the attention map has deviated under the quantization. To obtain the cover length ratio, first, sort the student and teacher's attention map in probability order. Then we can get the teacher map's Top-K tokens that receive the most attention in the teacher attention map. After that, we find out how many tokens we need to look at to get all the teacher map's Top-K tokens from the sorted student map. The number of tokens that each token has to look at for covering the teacher map's Top-K tokens is called cover length, and we normalize cover length by sequence length. We call this metric cover length ratio.

~\cite{ptq4vit} introduced pairwise ranking loss to keep the relative order of attention values. Pairwise ranking loss indicates how different the order of attention importance is between two attention maps. With any of the two tokens, if the order of the two tokens in the student map is different from that in the teacher map, we add the difference between the values of the two tokens to the loss. 

\begin{equation}
    \begin{aligned}
    &\mathcal{L}_{ranking}^{h} \\
    &=\sum_{i=1}^{n-1}\sum_{j=i+1}^{n}\Phi(({AM}_{i}^{S}-{AM}_{j}^{S})\\
    &\quad \quad \quad \quad \quad \quad \quad \cdot sign(AM_{i}^{T}-AM_{j}^{T})), \\
    &\mathcal{L}_{ranking}=\sum_{h=1}^{N_H} \mathcal{L}_{ranking}^{h}, \\
    \end{aligned}
\end{equation}
where $h$ is an index of attention head.
%\sum_{k=1}^{N_H}

\subsection{Visualization of Attention Map}
\label{appen:vis-map}
\begin{figure}
\begin{center}
\centerline{\includegraphics[width=0.7\columnwidth]{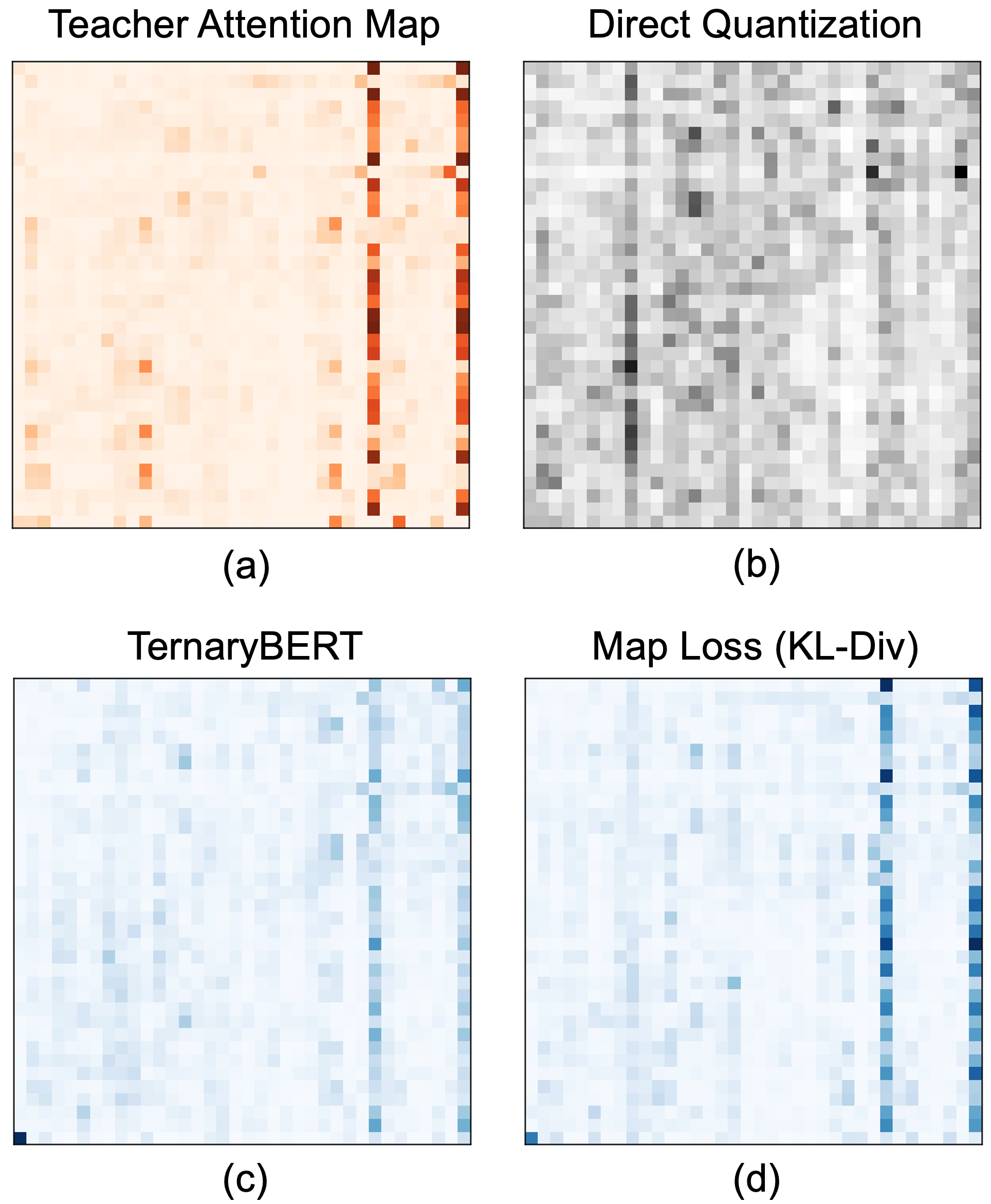}}
\caption{Visualization of self-attention map with BERT-Base over RTE Task (a) Teacher self-attention map (b) After quantization without QAT (c) After TernayBERT KD-QAT (d) Attention-map loss}
\label{fig:attention_map}
\end{center}
\vskip-0.3in
\end{figure}

We compare the self-attention map of fine-tuned full-precision BERT-base model and the quantized model on the RTE task.
Fig.~\ref{fig:attention_map} shows self-attention maps from the 3rd Transformer layer (8th head) of BERT-Base. Note that applying quantization distorts the self-attention map of the teacher severely. As shown in the figure, attention-map loss successfully recovers the self-attention map of the teacher, whereas TernaryBERT fails to capture some distinctive features. 

\subsection{Unified Attention-Map and Output Loss Exploration}
\label{appen:scale-mix_exploring}
As mentioned in Sec.~\ref{subsec:scaled-mix}, we explore the effectiveness of unifying the attention map and output losses for QAT. We conducted the experiments by fixing one of two losses, attention-map, and output loss, and changing the mixing parameter $\gamma$ for another loss according to Eq.~\ref{eq:mixed_scale}. Table ~\ref{tab:mixed_scale_base} and \ref{tab:mixed_scale_large} show the best results changing mixing parameter $\gamma$ in BERT-Base and BERT-Large. Overall, maintaining attention-map loss (case $SM_{1}$) perform better than opposite cases (case $SM_{2}$).
To show the degree of effect according to mixing parameter $\gamma$, we summarize the $\gamma$ showing the best result for each case in Table~\ref{tab:mixed_scale_gamma}. The table shows that every task has its own score favorable mixing parameter $\gamma$. These trends can be connected to the task-dependent attention characteristics in Sec~\ref{subsec:task-dependent}.

\begin{table}[H]
\centering
\resizebox{0.8\linewidth}{!}{
\begin{tabular}{c|cccc}
\toprule
GLUE & \multicolumn{2}{c}{BERT-Base} & \multicolumn{2}{c}{BERT-Large} \\
Task & $SM_1$ & $SM_2$ & $SM_1$ & $SM_2$ \\
\midrule
RTE   & 0.6  & 0.4   & 0.1  & 0.9  \\
CoLA  & 0.4  & 0.2   & 0.1  & 0.5  \\
STS-B & 0.7  & 0.2   & 0.3  & 0.9  \\
MRPC & 0.1  & 0.1   & 0.1  & 0.9  \\
SST-2 & 0.2  & 0.7   & 0.3  & 0.1  \\
QNLI & 0.8  & 0.8   & 0.8  & 0.6  \\
MNLI & 0.5  & 0.1   & 0.8  & 0.7  \\
QQP & 0.1  & 0.1   & 0.5  & 0.6  \\
\bottomrule
\end{tabular}}
\caption{Results of Map+Output mixing parameter $\gamma$ exploration}
\label{tab:mixed_scale_gamma}
\end{table}

\begin{table*}
\centering
\resizebox{1\linewidth}{!}{

\begin{tabular}{clllllllll}
\toprule
GLUE   Task & RTE            & CoLA           & STS-B          & MRPC           & SST-2          & QNLI           & MNLI           & QQP            & AVG   \\
\midrule
Full-Prec      & 73.28 & 58.04 & 89.24          & 87.77 & 92.09          & 91.32          & 84.37          & 89.30          & 83.39 \\
\midrule
$SM_{1}$     & \textbf{71.68} {\color{gray}\scriptsize±1.19} & \textbf{50.50} {\color{gray}\scriptsize±0.45}& \textbf{87.73} {\color{gray}\scriptsize±0.16} & 88.18 {\color{gray}\scriptsize±0.53}& \textbf{92.39} {\color{gray}\scriptsize±0.18} & 90.90 {\color{gray}\scriptsize±0.14}& \textbf{84.33} {\color{gray}\scriptsize±0.06}& \textbf{89.28} {\color{gray}\scriptsize±0.10}& 81.87 \\

$SM_{2}$  & 71.48 {\color{gray}\scriptsize±0.96} & 50.10 {\color{gray}\scriptsize±1.02} & 87.71 {\color{gray}\scriptsize±0.09} & \textbf{88.22} {\color{gray}\scriptsize±0.39}& 92.32 {\color{gray}\scriptsize±0.11} & \textbf{90.91} {\color{gray}\scriptsize±0.07} & 84.24 {\color{gray}\scriptsize±0.04} & 89.22 {\color{gray}\scriptsize±0.08} & 81.83 \\
\bottomrule
\end{tabular}}
\caption{BERT-Base Map+Output performance results on GLUE benchmark. Small dataset (under 10k) tasks are repeated 5 times; the others are repeated 3 times.}
\label{tab:mixed_scale_base}
\end{table*}

\begin{table*}
\centering
\resizebox{1\linewidth}{!}{
\begin{tabular}{clllllllll}
\toprule
GLUE Task & RTE            & CoLA           & STS-B          & MRPC           & SST-2          & QNLI           & MNLI           & QQP            & AVG     \\
\midrule
Full-Prec         & 70.39 & 60.31 & 89.83          & 88.43 & 92.32          & 92.29          & 86.49          & 89.55          & 83.70   \\
\midrule
$SM_{1}$    & \textbf{68.83} {\color{gray}\scriptsize±1.45} & \textbf{54.69} {\color{gray}\scriptsize±1.08}& 88.85 {\color{gray}\scriptsize±0.15}& \textbf{88.64} {\color{gray}\scriptsize±0.79}& \textbf{92.31} {\color{gray}\scriptsize±0.11}& \textbf{92.16} {\color{gray}\scriptsize±0.15}& 86.32 {\color{gray}\scriptsize±0.03}& \textbf{89.48} {\color{gray}\scriptsize±0.05}& 82.64   \\
$SM_{2}$ & 68.23 {\color{gray}\scriptsize±1.06}& 54.44 {\color{gray}\scriptsize±0.46}& \textbf{88.93} {\color{gray}\scriptsize±0.08} & 88.24 {\color{gray}\scriptsize±0.26} & 92.20 {\color{gray}\scriptsize±0.30} & \textbf{92.16} {\color{gray}\scriptsize±0.07} & \textbf{86.36} {\color{gray}\scriptsize±0.06} & 89.45 {\color{gray}\scriptsize±0.03} & 82.46 \\
\bottomrule
\end{tabular}}
\caption{BERT-Large Map+Output performance results on GLUE benchmark. Small dataset (under 10k) tasks are repeated 5 times; the others are repeated 3 times.}
\label{tab:mixed_scale_large}
\end{table*}

\subsection{Experimental Setup}
\label{appen:expr_setting}

\textbf{Datasets}\\
We evaluate our method on all datasets of the GLUE benchmark~\cite{glue} for BERT, and two datasets of the KLUE~\cite{klue}, which is one of the datasets to evaluate the natural language understanding capability of Korean language models, and sentiment analysis NSMC\footnote{\url{https://github.com/e9t/nsmc}} datasets. Details are as follows.
\begin{enumerate}
\item \textbf{GLUE.} The General Language Understanding Evaluation is a collection of resources for training, evaluating, and analyzing natural language understanding systems.
\item \textbf{KLUE-TC.} The KLUE Topic Classification is a single sentence classification task, and it classifies which topic the input sentence belongs to among the 7 representative topics. We averaged the accuracy and F1 score as the metric.
\item \textbf{KLUE-STS.} The KLUE Semantic Textual Similarity is to measure the degree of semantic similarity between two Korean sentences. We averaged the Pearson Correlation Coefficient (PCC) and Spearman Correlation Coefficient (SCC) to measure the performance. 
\item \textbf{NSMC.} The NAVER Sentiment Movie Corpus has a collection of movie reviews scraped from NAVER Movies\footnote{\url{https://movie.naver.com/movie/point/af/list.naver}}, including an annotation of whether the evaluation of the movie is positive or negative.
\end{enumerate}

\noindent \textbf{Training Settings} \\
For evaluating our methods, we use batch size 16 for CoLA and 32 for other GLUE tasks. The learning rate starts from zero and gradually increases to 2e-5 during the warm-up stage and decays linearly to 2e-9 for 3 epochs. The dropout probability was always kept at 0.1. For an optimizer, we use BertAdam\footnote{\url{https://github.com/huggingface/transformers/blob/v0.6.2/pytorch_pretrained_bert/optimization.py}}, which is a variant of Adam.
For ULM-Large, we train for 10 epochs using AdamW~\cite{adamw} for KLUE-TC and Adafactor~\cite{adafactor} for KLUE-STS and NSMC. We empirically find the best hyperparameter setting for each task in the following choices:
\begin{itemize}
    \item \textbf{Batch size:} 16, 32, 64
    \item \textbf{Learning rate:} 1e-5, 2e-5, 5e-5
\end{itemize}

\end{document}